%% file: main.tex
\documentclass[10pt,twocolumn,letterpaper]{article}

\usepackage{cvpr}              

\input{preamble}

\usepackage{graphicx}
\usepackage{amsmath}
\usepackage{amssymb}
\usepackage{booktabs}
\usepackage{multirow}
\definecolor{cvprblue}{rgb}{0.21,0.49,0.74}
\usepackage[pagebackref,breaklinks,colorlinks,citecolor=cvprblue]{hyperref}

\title{T2M-X: Learning Expressive Text-to-Motion Generation from Partially Annotated Data}
\renewcommand{\thefootnote}{\fnsymbol{footnote}}  
\author{
Mingdian Liu\textsuperscript{1,3}\footnotemark[1] \quad Yilin Liu\textsuperscript{2,3} \quad Gurunandan Krishnan\textsuperscript{3} \quad Karl S Bayer\textsuperscript{3} \quad Bing Zhou\textsuperscript{3}\footnotemark[2]\\
$^1$Iowa State University \quad 
$^2$Pennsylvania State University \quad 
$^3$Snap Inc.\\
}
\begin{document}
\maketitle

\begin{abstract}
   The generation of humanoid animation from text prompts can have profound impact on animation production and AR/VR experiences. However, existing methods only generate body motion data, excluding facial expressions and hand movements. This limitation, primarily due to a lack of comprehensive whole-body motion dataset, inhibits their readiness for production use. Recent attempts to create such a dataset have resulted in either motion inconsistency among different body parts in the artificially augmented data or lower quality in the data extracted from RGB videos.
   In this work, we propose T2M-X, a two-stage method that learns expressive text-to-motion generation from partially annotated data. T2M-X trains three separate Vector Quantized Variational AutoEncoders (VQ-VAEs) for body, hand and face on respective high quality data sources to ensure high quality motion outputs, and a Multi-indexing Generative Pretrained Transformer (GPT) model with motion consistency loss for motion generation and coordination among different body parts. Our results show significant improvements over the baselines both quantitatively and qualitatively, demonstrating its robustness against the dataset limitations.
\end{abstract}

  
\section{Introduction}
\label{sec:intro}
Over the past several years, text-to-motion models have emerged as a widely accepted and highly prevalent paradigm in the realm of computational modeling. It finds profound applications across diverse fields such as animation production, gaming and robotics. Conventionally, new motions in animations are obtained through motion capture or handcrafted manually, which is a costly process. Thus, the ability to automatically translate text into meaningful motion data could offer a more time-efficient and cost-effective solution.
\begin{figure}
    \centering
    \includegraphics[height=2in]{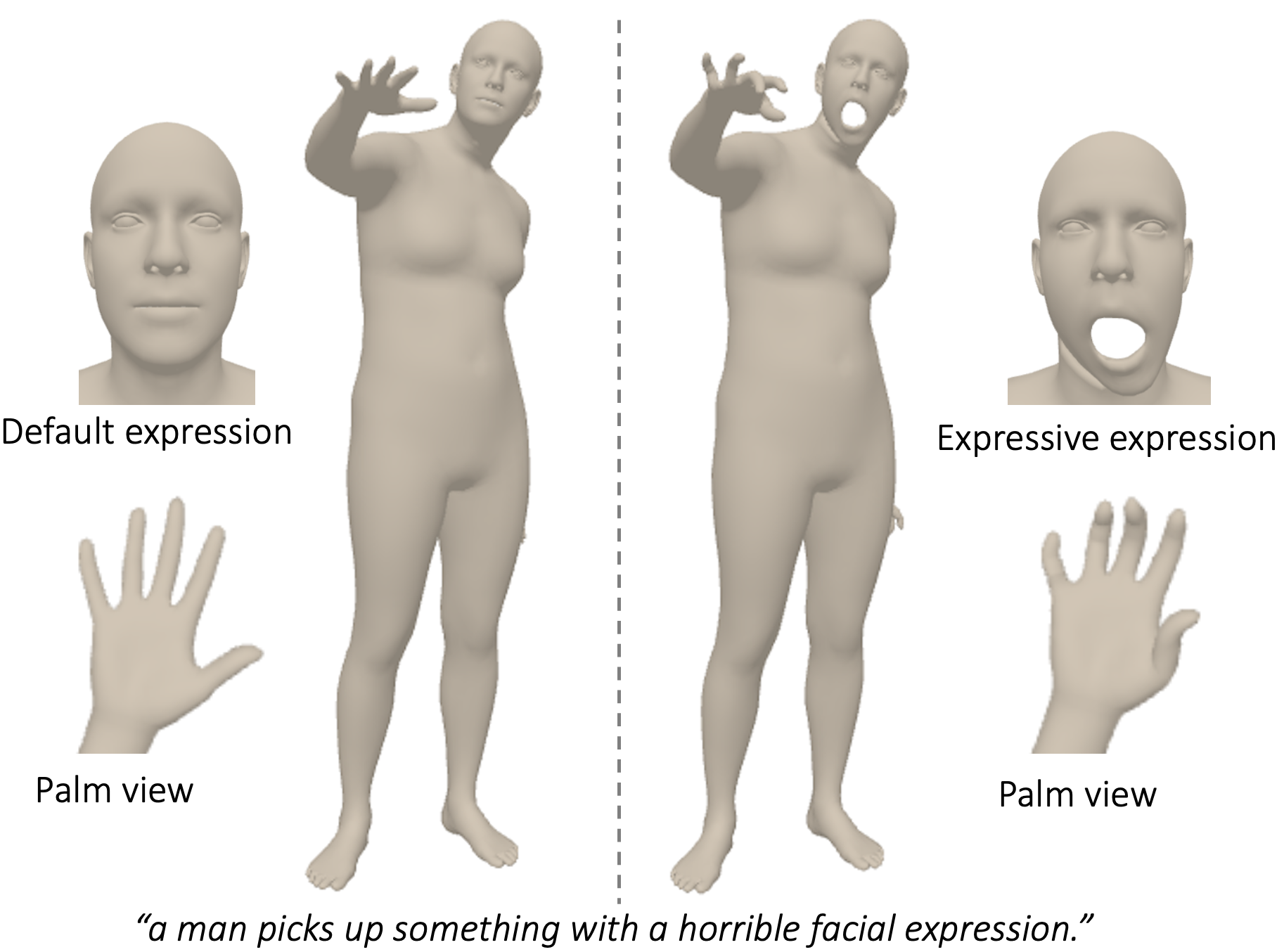}
    \caption{Difference between text to motion generation and text to expressive whole-body motion generation.}
    \label{fig:motion-x_diff}
\end{figure}
\footnotetext[1]{Work partially done during an internship at Snap Inc.}
\footnotetext[2]{Corresponding author}
\renewcommand{\thefootnote}{\Arabic{footnote}}
The task of generating motion conditioned on natural language presents a significant challenge due to the differing modalities of motion and text. This requires the model to accurately learn and execute a mapping from the linguistic space to the motion space. To achieve this goal, MotionCLIP \cite{tevet2022motionclip} leverages CLIP's \cite{radford2021learning} shared text-image latent space to improve out-of-distribution motion generation. TEMOS \cite{petrovich2022temos} combines a Transformer-based VAE and text encoder for non-autoregressive motion sequence generation. TEACH \cite{athanasiou2022teach} refines TEMOS, generating temporal motion compositions from text. T2M-GPT \cite{zhang2023t2m} uses the VQ-VAE~\cite{van2017neural} framework for competitive results. MotionDiffuse \cite{zhang2022motiondiffuse} pioneers the diffusion model for text-to-motion tasks with fine-grained instructions. MDM \cite{tevet2022human} employs a transformer-encoder backbone to adapt a diffusion-based model for human motion generation.
Nonetheless, despite these advancements, the generated motion only contains the body motion without facial expressions and hand motions, as shown in Figure~\ref{fig:motion-x_diff}. Current text-to-motion models, limited by the quality and quantity of available datasets, are yet to match the performance of state-of-the-art (SOTA) text-to-image models. This discrepancy is largely due to the substantial cost involved in collecting motion data and crafting corresponding text descriptions, a significant barrier to progress in the field.

\begin{figure}
    \centering
    \includegraphics[width=3.0in]{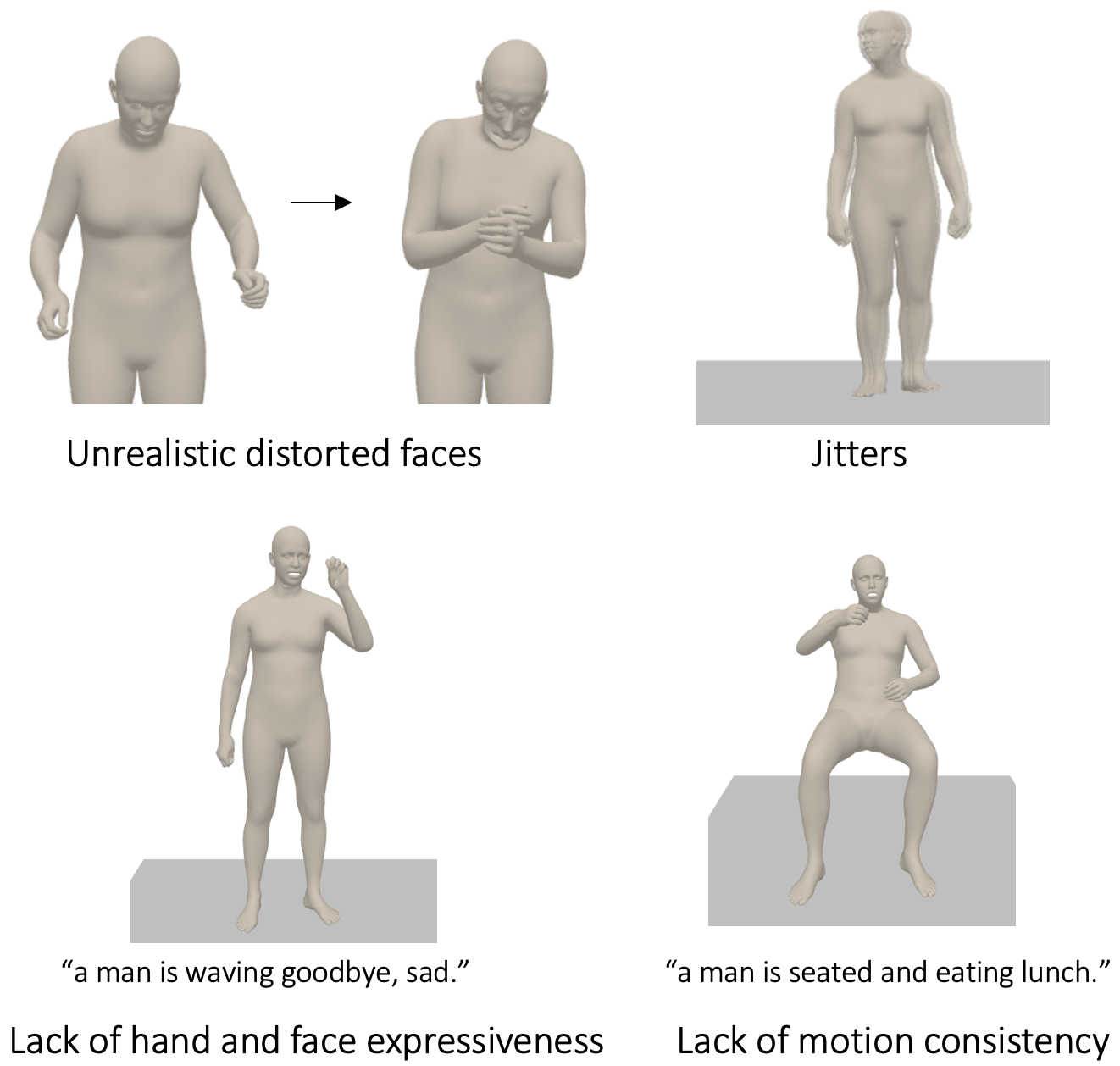}
    \caption{Limitations of existing whole-body motion datasets.}
    \label{fig:motion-x}
\end{figure}

In a recent endeavor towards comprehensive whole-body motion datasets, Motion-X~\cite{lin2023motionx} introduced a large-scale 3D dataset of expressive whole-body motions, each paired with a textual description. This was achieved by augmenting existing partially annotated datasets and extracting whole-body motions from videos. For instance, the HumanML3D~\cite{guo2022generating} dataset is enriched by the inclusion of facial expressions randomly selected from the high-quality BAUM~\cite{baum} facial expression dataset. Despite the superior motion data quality in both datasets, the augmented full-body motion exhibits a lack of congruence between body movements and facial motions. Moreover, the prevalent default hand pose in the HumanML3D data leads to similar behavior in the generated motion, limiting the expressiveness of hand motions. Indeed, Motion-X has pioneered a pipeline for capturing whole-body motion data from RGB videos, coupled with automatic text labelling via visual question answering models. Nevertheless, the quality of the extracted motion data seems to pale in comparison to that derived from professional devices. The motion exhibits jitters and facial distortions when the subject is not directly facing the camera or under occlusion, and the facial expressions appear somewhat subdued, likely due to the small face size in the video. Thus, the using such dataset for learning a whole-body motion generation model inevitably leads to several inherent limitations. These include jitter in the generated motion data, distorted facial expressions, inconsistency amongst different body parts, and a lack of expressiveness in hand and face motions, as illustrated in Figure~\ref{fig:motion-x}. Moreover, the text labels derived from vision models tend to be brief and unreliable, particularly in the realm of facial expression recognition. For instance, it's not uncommon to encounter a semantic description such as ``surprise expression and walking" paired with a facial expression label of ``sadness", which is labelled by a pretrained facial expression recognition model. Blindly assuming the dataset's completeness and learning from it may lead to perplexing outcomes.

\begin{table}
  \centering
  \resizebox{0.45\textwidth}{!}{\begin{tabular}{c c c c c c c c}
    \toprule
    \textbf{Datasets} & \textbf{Body} & \textbf{Hand} & \textbf{Face}  & \textbf{Text} & \textbf{Motion}\\
    \midrule
    HumanML3D~\cite{guo2022generating} & \checkmark &  &  & Precise & High\\
    GRAB~\cite{taheri2020grab}  & \checkmark & \checkmark &  & Basic & High\\
    BAUM~\cite{baum}  & \ & \ & \checkmark & Precise & High\\
    IDEA400~\cite{lin2023motionx}  & \checkmark & \checkmark & \checkmark & Basic & Average\\
    \bottomrule
  \end{tabular}}
  \caption{The quality of existing partially annotated datasets.}
  \label{tab:partial_data}
\end{table}

To achieve high-quality motion generation, the model would ideally learn from high-quality data. Regrettably, such datasets are scarce. Table~\ref{tab:partial_data} offers an overview of the major existing datasets, including their motion and text description quality. HumanML3D~\cite{guo2022generating}, BAUM~\cite{baum}, and GRAB~\cite{taheri2020grab} each contributes unique aspects of high-quality data, but they are all partially annotated, which means they lack either body, hand, or face data. The latest dataset, IDEA400~\cite{lin2023motionx}, a subset of Motion-X, stands out as it extracts whole-body motion from RGB videos, making it one of the few large-scale whole-body motion datasets. Despite Motion-X providing a complete and unified whole-body dataset by augmenting these partially annotated datasets, thus simplifying the generation of expressive whole-body motions, it unfortunately inherited the data artifacts and limitations. These include the lower motion quality and the visual artifacts and noise caused by the limitations of RGB-based 3D motion extraction.
Additionally, motion jitter becomes a more prominent issue in lower-quality datasets extracted from videos, which can be inherited by the generated motion outputs if not properly addressed.

In order to optimally utilize both high-quality partially annotated data and lower-quality whole-body motion data, we suggest a bifurcated approach to the task of motion generation. The complete process encompasses motion representation learning, accomplished via three distinct VQ-VAE expert models, and a multi-index GPT model for the generation of motion sequences for all three body parts based on text descriptions. The separate learning of VQ-VAE models from partially high-quality data sources guarantees the superior quality of the generated motions. Meanwhile, we employ a joint space consistency loss and facilitate the coordination of motions among the three body parts by training the GPT model with dynamic backpropagation. We only update the weights of the GPT base and the corresponding index branch when the respective body part data are available. Further, by training the GPT model across all datasets with consistency loss, we ensure the model has ample training data and learns from authentic whole-body motion data, as opposed to artificially augmented data.

Furthermore, we have gathered higher quality motion datasets, such as Mixamo~\cite{mixamo}, and converted them into the standard SMPL-X format to enhance the motion generation datasets. We've also suggested a motion jitter detection and mitigation strategy as a data preprocessing step to minimize jitter in lower-quality datasets. For datasets with very simplistic text descriptions, we utilize ChatGPT~\cite{chatgpt} to expand each short description into three full sentences in various ways. Results indicate that this significantly enhances the generalizability of our GPT model.

In summary, our contributions are as follows:
\begin{itemize}
    \item We created a unified, high-quality motion dataset, paired with detailed text descriptions, by standardizing data formats, and implementing both motion and text augmentations for enhanced text-to-motion generation.
     \item To maximize the use of high-quality partially annotated and lower-quality whole-body motion data, we propose a two-stage motion generation process which includes three VQ-VAE expert models trained on partially annotated data and a multi-indexing GPT model for motion sequence generation.
     \item We implemented a body part motion consistency loss in the joint space throughout the training process to achieve consistent body motions among all the body parts.
\end{itemize}

\section{Related Works}
\textbf{Text to Body Motion Generation.}
Generating 3D human motion from text description is a popular subtask of motion generation\cite{li2021ai, chen2023executing,mao2020history, pavllo2018quaternet, petrovich2021action, siyao2022bailando, starke2022deepphase, tang2022real, ghosh2021synthesis, chen2021choreomaster, aristidou2022rhythm, aliakbarian2020stochastic}. Early work such as Text2Action \cite{ahn2018text2action, lucas2022posegpt} uses an RNN to produce upper body motion from short texts. The KIT~\cite{plappert2016kit} motion-language dataset, introduced in 2016, paved the way for Language2Pose~\cite{ahuja2019language2pose}, which uses a curriculum learning strategy for a joint text and pose embedding, aiding motion sequence generation. Recently, BABEL \cite{punnakkal2021babel} and HumanML3D\cite{guo2022generating} have annotated the AMASS motion capture dataset with English labels, encouraging more multimodal model development.

More recent work, such as MotionCLIP \cite{tevet2022motionclip} capitalizes on the shared text-image latent space from CLIP \cite{radford2021learning} to enhance out-of-distribution generation beyond existing dataset constraints. TEMOS \cite{petrovich2022temos} trains a Transformer-based VAE for human motion in combination with an additional text encoder, allowing for the generation of various motion sequences based on a given textual description in a non-autoregressive manner. TEACH \cite{athanasiou2022teach} further refines TEMOS by enabling the generation of temporal motion compositions from a sequence of natural language descriptions. T2M-GPT \cite{zhang2023t2m} stands out by employing the classic VQ-VAE framework, reaching competitive results compared to previous methods. Apart from VAE architecture, the diffusion model has also been adapted for text-to-motion tasks. MotionDiffuse \cite{zhang2022motiondiffuse} pioneers this approach, focusing on fine-grained body part instructions and supporting synthesis of varied-length motions. Utilizing a transformer-encoder backbone, MDM \cite{tevet2022human} adapts a classifier-free, diffusion-based generative model for the human motion domain. 

While previous studies have been limited to generating body motion, our work extends this capability to hand gestures and facial expressions. We demonstrate how to utilize partially annotated datasets to concurrently produce body motion, hand gestures, and facial expressions.



\noindent\textbf{Consistency Learning.}
Generating data across multiple modalities can be considered a form of multi-task learning, where cross-task relations are adapted into multimodal generative models through consistency learning. This approach is particularly applied in handling partially annotated data, as exemplified in two previous papers. Lu \textit{et al}.\cite{lu2021taskology} introduce a method to utilize cross-task consistency between different task predictions on unlabeled data within a mediator dataset. This approach facilitates joint model learning for distributed training, employing consistency losses that reflect the coherence between adjacent video frames. Li \textit{et al}~\cite{li2022learning} extend \cite{lu2021taskology} to multi-task predictions from partially annotated data by developing an auxiliary network. This network learns a unified task-space for each task pair, optimizing computational efficiency through shared parameters across different mappings and task-pair dependent outputs.

In our work, we apply consistency learning to address partially annotated data within a generative model. Different from \cite{lu2021taskology} and ~\cite{li2022learning}, our approach involves learning a single joint space specifically for motion data across three modalities. Additionally, we design a novel feature extractor based on the GRU layer to take the previous motion into account which is essential for sequence data generation.

\section{Method}
\begin{figure*}
    \centering
    \includegraphics[width=6.0in]{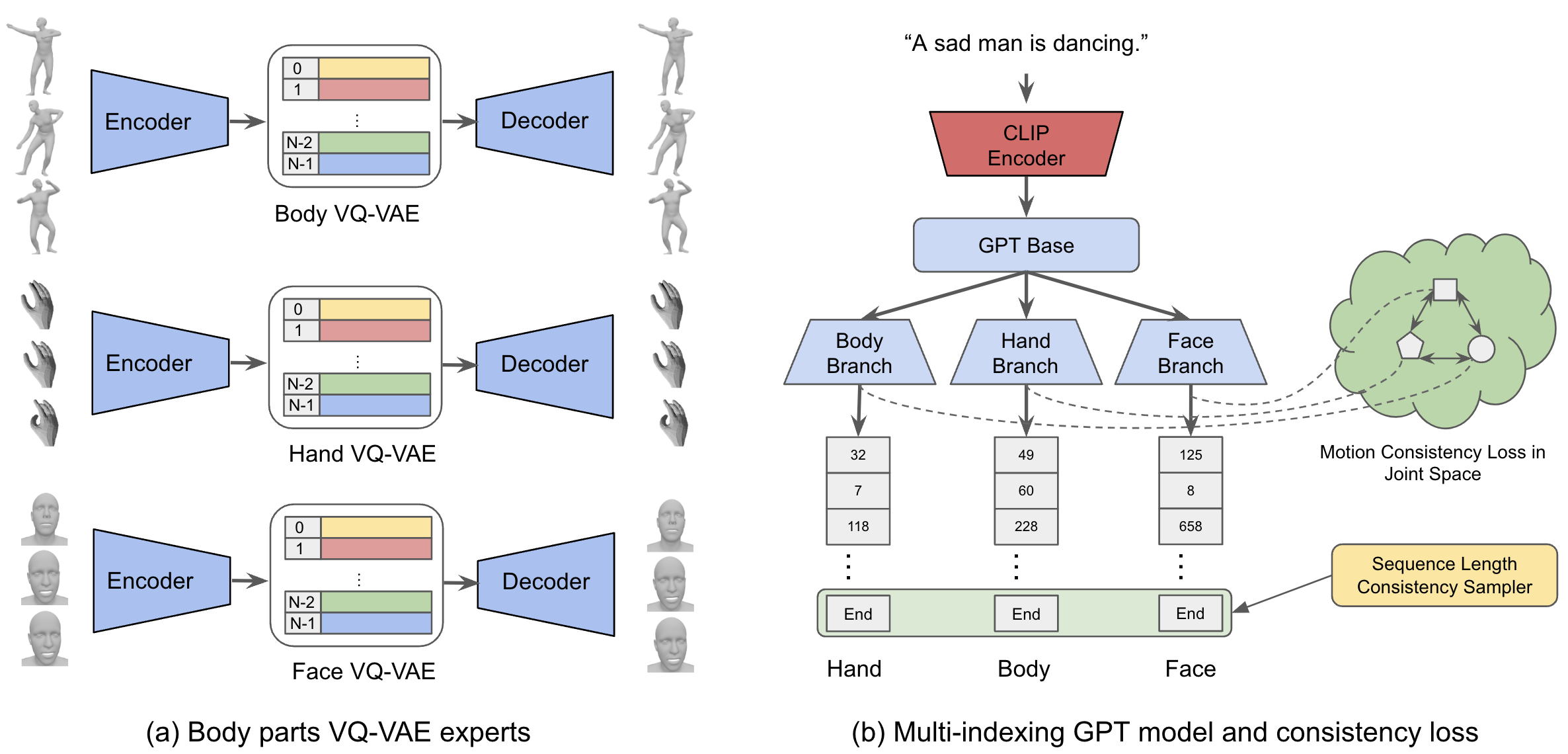}
    \caption{The architecture of VQ-VAE experts for body, hand, and face motion token generation, multi-indexing GPT for coordination, and the joint space for consistency learning.}
    \label{fig:network}
\end{figure*}

Our model transforms text into expressive whole-body motion data in SMPL-X format. As depicted in Figure~\ref{fig:network}, it includes three VQ-VAE models, each decoding different body part movements into index sequences. A multi-indexing GPT model predicts whole-body motion sequences from text, which VQ-VAE decoders then convert into motion data. To ensure consistency across motion modalities, a motion consistency loss is utilized during training.

\subsection{VQ-VAE Experts}
Since the model is required to learn form the partially annotated data, we try to tackle our task with the VQ-VAE experts for three motion modalities--body, hand, and face. The expert models are trained on the motion data of high quality to project the original data into the codebook in sparse representation, as illustrated in Figure~\ref{fig:network}. Specifically, a series of human motion can be presented as \(M^p \in \mathbb{R}^{T\times d_m^p}\), where \(M^p\) is the partial motion data for body (\(p=1\)), hand (\(p=2\)), face (\(p=3\)), \(T\) is the number of frames and \(d_m^p\) is the data dimension of the motion modality. We aim to leverage an encoder-decoder of several 1-D convuluaitonal layers to compress the whole motion dataset of each modality into a learnable codebook \(C^p \in \mathbb{R}^{K^p\times d_c^p}\), where \(K^p\) and \(d_c^p\) are the length and dimension of the coodebook, and \(p\) is the modality index. As for motion \(M^p \in \mathbb{R}^{T\times d_m^p}\), the latent vector extracted from the encoder is denoted as \(z^p \in \mathbb{R}^{T^{\prime}\times d_c^p}\), where \(T^{\prime}=T/l\) and \(l\) is the downsampling rate of the encoder in temporal domain, and \(d_c^p\) is the vector dimension same as that of the codebook. To quantize the \(i\)-th latent vector \(z_i^p\), a closest vector in the codebook is found out to represent it as follows: 
\begin{equation}
  \hat{z}_i^p=\arg \min_{c_k^p\in C^p} ||z_i^p-c_k^p||_2 \in \mathbb{R}^{d_c^p}
  \label{eq:vqvae-1}
\end{equation}
and the index of the vector in the codebook is record as the motion token \(t_{m,i}^p \in \mathbb{R}^1\). In general, the motion \(M^p\) is projected into \(z^p\) by the encoder, quantized as \(\hat{z}^p\) by the codebook, and also represented by the motion token \(t_m^p\). The quantized vector \(\hat{z}^p\) will be used by the decoder, which operates with an architecture opposite to that of the encoder, to reconstruct the motion data \(\hat{M}^p\), and \(t_m^p\) will be fed into the multi-indexing GPT for model training.

We follow a similar strategy in \cite{zhang2023t2m} to optimize the VQ-VAE model for each motion modality. The final loss consists of a standard reconstruction loss for data reconstruction, a codebook alignment loss to update the codebook vector, and a commitment loss to penalize the gap between the output vector from encoder and the closest vector in the coodebook. The euqation is written as follows:
\begin{equation}
  \mathcal{L}_{vq}^p = \mathcal{L}_{re}^p + ||sg[z^p]-\hat{z}^p||_2 + \beta||z^p-sg[\hat{z}^p]||_2
  \label{eq:vqvae-2}
\end{equation}
where \(\beta\) is a hyper-parameter for the commitment loss and \(sg\) is the symbol for stop-gradient operator. For the reconstruction loss, we employ the smooth $\mathcal{L}1$ loss for all the motion modalities and an additional loss on the body motion velocity \(V(M^p)\) to mitigate the jittering issue\cite{siyao2022bailando}. Specifically, 
\begin{equation}
  \mathcal{L}_{re}^p=||M^p-\hat{M}^p||_1+\alpha||V(M^p)-V(\hat{M}^p)||
  \label{eq:vqvae-3}
\end{equation}
where \(\alpha=0\) for hand and face motion, and the optimal \(\alpha=0.5\) in \cite{zhang2023t2m} for body motion.

\subsection{Multi-indexing GPT} \label{method-GPT}
To train the generative model on the partially annotated motion data, we apply a GPT model with base and branch architecture to enable the partial back-propogation accross the model. As shown in Figure~\ref{fig:network}, the base model is used to process the text condition and the previous body motion tokens while the branch models is for predicting the next token indexes for body, hand, and face modalities. The GPT model functions simultaneously as a generator and coordinator of token indices in VQ-VAE experts, a process called multi-indexing GPT in this paper. 

Since human motion are mainly based on the body motion, and SMPL-X body part includes the wrist and head joints bridging to hand motion and facial expression, it is reasonable to condition the future body motion, hand motion, and facial expression on the text description and prior body motion. Specifically, given a text prompt \(c\) and three token sequences from the VQ-VAE experts \(\hat{z}^1\), \(\hat{z}^2\), and \(\hat{z}^3\), the GPT model is trained to align the predicted probability distribution \(\hat{P}_t^p = P(\hat{z}_{t}^p|c, \hat{z}_{1}^1,...,\hat{z}_{t-1}^1)\) to the ground truth distribution for body, hand, and face motions. We also add a special \(End\) token to the end of motion tokens to indicate the stop of human motion. The loss function for the multi-indexing GPT model consists of three cross entropy losses which can be written as follows:
\begin{equation}
\begin{split}
  \mathcal{L}_{GPT} & =\frac{1}{T+1} \sum_{t=1}^{T+1}\sum_{n=1}^{N_1+1} [-P_{tn}^1 \log \hat{P}_{tn}^1] \\ 
  & + \eta_1 \cdot \frac{1}{T+1} \sum_{t=1}^{T+1}\sum_{n=1}^{N_2+1} [-P_{tn}^2 \log \hat{P}_{tn}^2] \\ 
  & + \eta_2 \cdot \frac{1}{T+1} \sum_{t=1}^{T+1}\sum_{n=1}^{N_3+1} [-P_{tn}^3 \log \hat{P}_{tn}^3]
  \end{split}
  \label{eq:gpt-loss}
\end{equation}
where \(T\) is the length of motion token, \(N\) is the number of classes for the motion tokens, and \(\eta_1\) and \(\eta_2\) are the hyper-parameters for adjusting the loss weight of hand motion and facial expression. 

\noindent \textbf{Architecture.} Same as \cite{zhang2023t2m}\cite{tevet2022motionclip}\cite{lin2023motionx}, we feed the text prompt into the CLIP~\cite{clip} text encoder as the starter for the motion generation. The previous motion tokens are added to the text embedding to predict the future motions. It is noted that to leverage the token relevance learned in body VQ-VAE model, token embedding from the fixed codebook is input to GPT base as previous body motions, which is different from the simple \(nn.Embedding\) setting in \cite{zhang2023t2m}. In addition, the concatenated text embedding and motion embedding are fed into Transformer layers for feature extraction. Since the multi-indexing GPT works in an autoregressive manner, the causal self-attention \cite{vaswani2017attention} is utilized in the transformer layer to establish the computational dependency among sequential elements of data. The formula is written as follows:
\begin{equation}
  \text{Attention}(Q,K,V)=\text{Softmax}(\frac{QK^T \times M}{\sqrt{d_k}})V
  \label{eq:casual}
\end{equation}
where \(Q\), \(K\), \(V\), denote the query, key and value from input, and \(M\) is the causal mask to only allow the previous data to compute the current state of motion tokens. The extracted feature from GPT base is fed into the GPT branch parts to generate the next motion tokens for body, hand, and face modalities. The GPT branches share the same configuration for transformer layers as GPT base except a linear layer is attached to enable the calculation of Softmax function in equation \ref{eq:gpt-loss}.

\noindent\textbf{Training and Inference.} During the model training stage, it is first necessary to process data with partial annotations to enable effective backpropagation throughout the entire GPT model. We randomize the training data into batches in a way that ensures each batch contains at least one sample from every data modality. Two masks are also stored in the batch data to indicate the presence or absence of hand motion and face motion in each data sample. Based on these masks, we compute the average loss for each modality in equation \ref{eq:gpt-loss}, only for the data annotated with that specific modality. This setup ensures that the model weights in all GPT branches are updated with each batch of data, thereby enhancing the stability of the training process. In the inference stage, the motion stop token \(End\) may appear earlier in hand or face motion than in body motion. To maintain consistency in sequence length across different data modalities, a sequence length consistency sampler is employed. This sampler replaces the stop token from predicted hand/face sequences with the token of the second-highest probability when the stop token appears prematurely. In this way, the complete body motion, hand motion, and facial expression can be generated at the same time conditioned on the text prompt.
\subsection{Consistency Learning in Joint Space}
In the GPT model outlined in Section \ref{method-GPT}, there's a challenge in ensuring consistency among the generated motion tokens for different modalities, as the model doesn't fully capture the joint distribution of all motion types across the dataset. To address this, we create a joint space for body motion, hand motion, and facial expressions  as illustrated in Figure~\ref{fig:network}, to ensure the coherence in generated motions. We adopt the same GRU layer used in \cite{guo2022generating} as our feature extractor for each modality, mapping the motion tokens into corresponding embedding spaces. For processing these tokens, we apply a one-hot encoding followed by concatenation with the classification probabilities from the GPT branch. This approach not only ensures the GRU layer effectively learns the feature mappings across the entire motion sequence, but also enables the backpropagation of the loss in joint space to the GPT model for weight updating. In general, for a ground truth motion data of T tokens, the consistency loss and the final loss can be written as follows:
\begin{equation}
  \mathcal{L}_{Consist}^t=\lambda_1\mathcal{L}_{cl}(e_t^1, e_t^2)+\lambda_2\mathcal{L}_{cl}(e_t^1, e_t^3)+\lambda_3\mathcal{L}_{cl}(e_t^2, e_t^3)
  \label{eq:gpt-loss}
\end{equation}
\begin{equation}
  \mathcal{L}_{Final}=\mathcal{L}_{GPT}+\frac{1}{T+1} \sum_{t=1}^{T+1}\mathcal{L}_{Consist}^t
  \label{eq:gpt-loss}
\end{equation}
where \(e_t^1, e_t^2, e_t^3\) are the features extracted from the GRU layers for body, hand and face, \(\mathcal{L}_{cl}\) is the contrastive loss, and \(\lambda_1, \lambda_2, \lambda_3\) are hyper-parameters for the modality matching losses.

\subsection{Motion Jitter Mitigation}
We address motion jitter in lower quality dataset with multiple strategies. Firstly, we measure the jitters in the motion data by quantifying the high frequency component and apply a low-pass filter to mitigate them. Secondly, we recognize that root joint jitters influence all body joint positions, but the local joint rotations remain unaffected by propagated error. Existing works using the HumanML3D pose representation all involve a joint rotation reverse process from predicted joint positions using SMPLify~\cite{SMPL-X:2019} model in the visualization, which inevitably includes jitters from the predicted joint positions. This problem is subtle when the model is trained on jitter free datasets, but becomes prominent with lower quality data. To counter this, we suggest a novel pose representation in Section~\ref{sec:dataset} that incorporates both SMPL-X joint rotations and positions. The model predicts both during inference. For visualization, we selectively animate the upper body using the predicted rotations for a smoother outcome and use the reversed rotations from joint positions for the legs to ensure consistency between foot positions and root translations, thereby minimizing foot sliding. 

\section{Dataset Construction and Formalization}\label{sec:dataset}

\textbf{Dataset Construction.} We have constructed a new text-to-motion dataset based on Mixamo~\cite{mixamo}, characterized by high-quality body/hand motion and text descriptions. For GRAB~\cite{taheri2020grab} data, we ensured precision in text labels describing each hand and body movement through manual labeling, offering more details than the simplistic text descriptions in Motion-X. In conjunction with body motion dataset HumanML3D~\cite{guo2022generating} and face expression dataset BAUM~\cite{baum}, all four datasets we've employed are of superior quality and feature partial annotations. In terms of whole-body motion datasets, we quantified and smoothed out jitters in motion datasets like IDEA400~\cite{lin2023motionx}, and enhanced text descriptions by generating three comprehensive sentences from simple description labels. Finally, we implemented mirror augmentation for all motion data and rephrased and augmented mirror text descriptions.
All datasets are converted to SMPL-X~\cite{SMPL-X:2019} format to be consumed in model training. 
Table~\ref{tab: dataset-summary} shows the statistics of the proposed dataset. In total, our dataset has 61.4K of sequence, 16.6M of frame, 115.3h of human motion, 3.96M of text words.
\begin{table}
  \centering
  \resizebox{0.45\textwidth}{!}{\begin{tabular}{c c c c c c c c }
    \toprule
    \textbf{Dataset} & \textbf{Sequence} & \textbf{Frame} & \textbf{Duration} & \textbf{Word} & \textbf{Modality}\\
    \midrule
    HumanML3D & 29.2K & 7.1M & 57.2h & 2.02M & B \\
    BAUM  & 1.4K & 0.2M & 1.9h & - & F\\
    Mixamo  & 4.5K & 1.1M & 2.6h & 0.26M & B,H\\
    GRAB  & 2.7K & 3.2M & 7.4h & 0.49M & B,H\\
    IDEA400  & 25.0K & 5.2M & 48.1h & 1.19M & B,H,F\\
    \midrule
    Total  & 61.4K & 16.6M & 115.3h  & 3.96M & - \\
    \bottomrule
  \end{tabular}}
  \caption{The statistics of datasets after mirroring augmentation for body motion and hand motion. B, H, F are body, hand, and face.}
  \vspace{-10pt}
  \label{tab: dataset-summary}
\end{table}

\noindent\textbf{Dataset Formalization.} As we incorporated new datasets such as Mixamo, we need to address the inconsistent formats across our data sources. 
It's important to note that some datasets necessitate manual offset correction due to varying default poses. For instance, we manually fixed the hand pose offsets in Mixamo dataset so that it's fully compatible with SMPL-X format. 
Addressing these discrepancies manually is essential; otherwise, they could lead to inconsistencies during model training and make model convergence more difficult.

\noindent\textbf{Pose Representation.} 
We convert all the rotations into 6D rotation vectors as it has been shown to produce smoother motion results~\cite{6Dcontinuity}. In general, a complete whole-body pose \(\boldsymbol{p}\) in our work is defined as a tuple of (\(\boldsymbol{r}, \boldsymbol{b}^p, \boldsymbol{b}^v, \boldsymbol{b}^r, \boldsymbol{c}^f, \boldsymbol{h}, \boldsymbol{j}, \boldsymbol{f}\)), where \(\boldsymbol{r} \in \mathbb{R}^4\) is the vector for root joint defined in HumanML3D format~\cite{guo2022generating}; \(\boldsymbol{b}^p \in \mathbb{R}^{3\times 21}\), \(\boldsymbol{b}^v \in \mathbb{R}^{3\times 21}\), and \(\boldsymbol{b}^r \in \mathbb{R}^{6\times 21}\) are the local body joint positions, velocities and rotations in root joint space; \(\boldsymbol{c}^f \in \mathbb{R}^{4}\) is binary features for the foot ground contacts defined by~\cite{guo2022generating}; \(\boldsymbol{h} \in \mathbb{R}^{6\times 30}\) is the hand joint rotations in wrist joint spaces; \(\boldsymbol{j} \in \mathbb{R}^{6}\) is the jaw joint rotation for face motion and \(\boldsymbol{f} \in \mathbb{R}^{50}\) is the facial expression feature in SMPL-X format.

In our experiment, we found the commonly used HumanML3D body pose representation suffers rotation information loss for head and wrist joints because it approximates the joint rotations with SMPLify model~\cite{SMPL-X:2019} from the predicted joint positions, which may cause body part rotation artifacts in some cases.
Instead, we animate the upper body using predicted SMPL-X rotations to maintain smooth motions, and the lower body using reversed rotations from predicted joint positions to minimize foot sliding. 

\section{Experiments}
\subsection{Experimental Settings}
\textbf{Datasets.} We randomly shuffle the whole dataset we curated and split it into training, validation and test datasets, each has 80\%, 10\% and 10\% data samples. The total motion length for the training set is 92.3 hours and it contains 49,100 textual descriptions. We downsample all motion datasets to 30 frames-per-second. The VQ-VAE experts are trained on different dataset combinations which contain the respective high quality motion data, as indicated in Table ~\ref{tab: dataset-summary}.

\begin{table*}[t!]
  \centering
  \resizebox{0.95\textwidth}{!}{
  \begin{tabular}{l c c c c c c c}
    \toprule
    \multirow{2}{*}{Methods} & \multicolumn{3}{c}{R-Precision $\uparrow$}  & \multirow{2}{*}{FID $\downarrow$} & \multirow{2}{*}{MM Dist $\downarrow$} & \multirow{2}{*}{Diversity$\uparrow$} & \multirow{2}{*}{MModality $\uparrow$} \\\cline{2-4}
    
    & Top-1 & Top-2 & Top-3 & & & & \\

    \midrule
    Real & \(0.474^{\pm0.003}\) & \(0.684^{\pm0.006}\) & \(0.778^{\pm0.008}\) & \(0.000^{\pm0.000}\) & \(3.121^{\pm0.004}\) & \(11.392^{\pm0.188}\) & - \\
    VQVAE  & \(0.385^{\pm0.004}\) & \(0.627^{\pm0.006}\) & \(0.736^{\pm0.005}\) & \(0.640^{\pm0.015}\) & \(3.812^{\pm0.036}\) & \(11.330^{\pm0.173}\) & -  \\
    \midrule
    T2M-GPT  & \(0.288^{\pm0.003}\) & \(0.554^{\pm0.007}\) & \(0.654^{\pm0.008}\) & \(2.543^{\pm0.008}\) & \(4.784^{\pm0.046}\) & \boldmath{\(10.747^{\pm0.239}\)} & \(1.753^{\pm0.013}\) \\
    T2M-X w/o \(cl\) & \(0.297^{\pm0.004}\) & \(0.572^{\pm0.011}\) & \(0.663^{\pm0.014}\) & \(2.273^{\pm0.009}\) & \(4.529^{\pm0.029}\) & \(10.357^{\pm0.188}\) & \(1.929^{\pm0.035}\) \\
    T2M-X w/ \(cl\) & \boldmath{\(0.327^{\pm0.006}\)} & \boldmath{\(0.591^{\pm0.009}\)} & \boldmath{\(0.725^{\pm0.009}\)} & \boldmath{\(1.753^{\pm0.010}\)} & \boldmath{\(4.028^{\pm0.039}\)} & \(10.719^{\pm0.153}\) & \boldmath{\(2.265^{\pm0.024}\)} \\
    \bottomrule
  \end{tabular}}
  \caption{The evaluation results for text to body motion generation. \(cl\) denotes consistency loss}
  \vspace{-10pt}
  \label{tab: metrics}
\end{table*}

\noindent\textbf{Implementation Details.} We reuse the most of the optimal configuration in~\cite{zhang2023t2m} except a few differences. Specifically, for the expert model, we choose a codebook size of \(512\times 512\) and downsampling rate of 4 for all the three VQ-VAE models. We employ the AdamW optimizer of a default setting with a batch size of 256 and a learning rate of 1e$-4$ to optimize our VQ-VAE experts.

Since the motion token have different lengths, we use a maximum token length of 128 (17s for raw motion data) for training GPT model with a padding strategy. We employ 9 transformer layers for GPT base and each GPT branch with the hidden dimension of 512 and 16 heads. We also employ the weight on hand and face motion for next motion token prediction, and the weight on body-hand matching, body-face matching, and hand-face matching for consistency learning. The whole multi-indexing GPT model is optimized with a batch size of 256 by the AdamW optimizer with \([\beta_1, \beta_2] = [0.5, 0.99]\) and a step learning rate proposed in~\cite{zhang2023t2m}. We train all models on a single A100 80GB GPU. It takes approximately 16 hours and 48 hours for the VQ-VAE experts and multi-indexing GPT model respectively to reach convergence. For the following experiments, we choose the optimal hyper-parameter for our model while providing the experiments for hyper-parameter optimization in supplementary materials.

\begin{table*}[h]
  \centering
  \resizebox{0.95\textwidth}{!}{
  \begin{tabular}{c|c c c c|c c c c}
    \toprule
    \multirow{2}{*}{Train Set} & \multicolumn{4}{c}{HumanML3D (Test)} \vline & \multicolumn{4}{c}{T2M-X Dataset (Test)} \\
    & R-Precision $\uparrow$ & FID $\downarrow$ & Diversity $\uparrow$ & MModality $\uparrow$ & R-Precision $\uparrow$ & FID $\downarrow$ & Diversity $\uparrow$ & MModality $\uparrow$\\       
    \midrule
    Real & \(0.731^{\pm0.003}\) & \(0.000^{\pm0.000}\) & \(10.422^{\pm0.099}\) & - & \(0.778^{\pm0.008}\) & \(0.000^{\pm0.000}\) & \(11.392^{\pm0.188}\) & - \\
    \midrule
    HumanML3D  & \(0.605^{\pm0.006}\) & \boldmath{\(1.277^{\pm0.071}\)} & \(10.182^{\pm0.084}\) & \(1.072^{\pm0.074}\) 
    & \(0.515^{\pm0.011}\) & \(8.184^{\pm0.156}\) & \(9.034^{\pm0.109}\) & \(0.826^{\pm0.016}\) \\
    T2M-X Dataset  & \boldmath{\(0.633^{\pm0.006}\)} & \(2.762^{\pm0.081}\) & \boldmath{\(10.820^{\pm0.133}\)} & \boldmath{\(1.344^{\pm0.064}\)} 
    & \boldmath{\(0.654^{\pm0.008}\)} & \boldmath{\(2.543^{\pm0.008}\)} & \boldmath{\(10.747^{\pm0.239}\)} & 
    \boldmath{\(1.753^{\pm0.013}\)} \\
    \bottomrule
  \end{tabular}}
  \caption{The evaluation results across HumanML3D and T2M-X dataset. We train T2M-GPT on the training set of HumanML3D and T2M-X, respectively, then evaluate this model on their test set.}
  \label{tab: cross-dataset}
\end{table*}

\begin{table*}
  \centering
  \resizebox{0.8\textwidth}{!}{\begin{tabular}{c c c c c c c}
    \toprule
    Modality Pair& \multirow{2}{*}{Methods} & \multicolumn{2}{c}{R-Precision $\uparrow$} & \multirow{2}{*}{MM Dist$\downarrow$} & \multicolumn{2}{c}{FID$\downarrow$}\\
    (a,b) & & b$\rightarrow$a & a$\rightarrow$b & & a & b\\
    \midrule
    \multirow{3}{*}{Body, Hand} & Real & \(0.733^{\pm0.002}\) & \(0.717^{\pm0.004}\) & \(3.514^{\pm0.013}\) & 
    \(0.000^{\pm0.000}\) & \(0.000^{\pm0.000}\) \\
    & w/o \(cl\)& \(0.620^{\pm0.014}\) & \(0.594^{\pm0.012}\) & \(5.726^{\pm0.085}\) & 
    \(2.339^{\pm0.021}\) & \(3.487^{\pm0.035}\) \\
    & w/ \(cl\) & \boldmath{\(0.701^{\pm0.005}\)} & \boldmath{\(0.659^{\pm0.008}\)} & \boldmath{\(3.836^{\pm0.020}\)} & 
    \boldmath{\(1.632^{\pm0.017}\)} & \boldmath{\(3.008^{\pm0.028}\)}\\
    \midrule
    \multirow{3}{*}{Body, Face} & Real & \(0.718^{\pm0.007}\) & \(0.682^{\pm0.005}\) & \(3.974^{\pm0.024}\) & 
    \(0.000^{\pm0.000}\) & \(0.000^{\pm0.000}\) \\
    & w/o \(cl\)& \(0.617^{\pm0.017}\) & \(0.572^{\pm0.056}\) & \(5.222^{\pm0.087}\) & 
    \boldmath{\(3.354^{\pm0.045}\)} & \(2.247^{\pm0.032}\) \\
    & w/ \(cl\) & \boldmath{\(0.704^{\pm0.042}\)} & \boldmath{\(0.612^{\pm0.036}\)} & \boldmath{\(4.457^{\pm0.056}\)} & 
    \(3.575^{\pm0.060}\) & \boldmath{\(1.755^{\pm0.023}\)} \\
    \midrule
    \multirow{3}{*}{Hand, Face} & Real & \(0.643^{\pm0.013}\) & \(0.631^{\pm0.009}\) & \(4.535^{\pm0.033}\) & 
    \(0.000^{\pm0.000}\) & \(0.000^{\pm0.000}\) \\
    & w/o \(cl\)& \(0.594^{\pm0.033}\) & \(0.537^{\pm0.026}\) & \(5.626^{\pm0.086}\) & 
    \(3.850^{\pm0.091}\) & \(2.247^{\pm0.032}\) \\
    & w/ \(cl\) & \boldmath{\(0.618^{\pm0.027}\)} & \boldmath{\(0.572^{\pm0.018}\)} & \boldmath{\(5.420^{\pm0.095}\)} & \boldmath{\(3.372^{\pm0.079}\)} & \boldmath{\(1.755^{\pm0.023}\)} \\
    \bottomrule
  \end{tabular}}
  \caption{The evaluation results for body-hand matching. \(cl\) denotes consistency loss.}
  \label{tab:body-hand-matching}
  \vspace{-10pt}
\end{table*}

\noindent\textbf{Evaluation Metrics.} In this paper, we employ the methodology of~\cite{guo2022generating} to evaluate the performance of text to motion generation. In particular, for text to motion generation, the text descriptions and generated motions are converted to embedding features by the pretrained network designed in~\cite{guo2022generating} to calculate five metrics as follows: \emph{Frechet Inception Distance (FID)}: the distribution distance between the real and generated motion on the extracted motion features; \emph{Diversity}: the average Euclidean distances of 300 pairs of motion features randomly sampled from a set; \emph{Multimodality}: the average Euclidean distances of 10 pairs of generated motion features from the same text description; \emph{Multimodal Distance}: the average Euclidean distance between each text feature and the generated motion feature from this text; \emph{R-Precision}: the average motion-to-text retrieval accuracy ranked by \emph{Multimodal Distance}. 

Meanwhile, we also create additional evaluation metrics for motion modality matching. We reuse the network setting in~\cite{guo2022generating} to train the feature extractors for the body, hand, and face motions on different dataset combinations. 32 text descriptions are randomly chosen from the test dataset to generate corresponding body, hand and face motions. For each modality-pair \emph{i.e.} body-hand, body-face, and hand-face, the generated motions are fed into the pretrained extractor to calculate the \emph{Multimodal Distance}, \emph{R-Precision}, and \emph{FID} in feature space. More details are provided in supplementary material due to the space limit.

\subsection{Quantitative Evaluation}
We evaluate our T2M-X model and dataset quantitatively and compare the results with the baseline state-of-the-art (SOTA) model and HumanML3D dataset.

\noindent\textbf{Text to Body Motion Generation.}
Since the body motion serves as the pivotal of human motion, we firstly evaluate our model on text to body motion generation against the SOTA T2M-GPT~\cite{zhang2023t2m}. The models are trained on our T2M-X dataset and share the same body motion VQ-VAE expert. The multi-indexing GPT model is explored with and without consistency learning for comparison. We compare the body motion component of the generated outputs as the T2M-GPT does not output hand and face motions. 
Table~\ref{tab: metrics} shows the comparison results.
The baseline T2M-GPT model falls short in \emph{R-Precision}, \emph{FID}, \emph{MM-Dist}, and \emph{MModality}. This limitation arises because our dataset's text descriptions encompass hand and face movements, which cannot be adequately represented through body motion alone. Our multi-indexing GPT without consistency learning outperforms the baseline in almost all the metrics which demonstrates our model is better at capturing the textual semantics. This is mainly because most of the textual descriptions in our dataset contains detailed description of the hands and face motion, which can be efficiently mapped to our model output, while such motions are not captured by T2M-GPT. 
The incorporation of consistency learning enables our model to synchronize body motion more harmoniously with the text descriptions related to hand and face movements, which brings a significant boost to all evaluation metrics. This leads to the generation of more intricate and diverse body motions, which can fully encapsulate the meaning of the text descriptions for body, hand, and face movements.

\noindent\textbf{Cross Dataset Evaluation.} To demonstrate the advantages of our proposed T2M-X dataset for the task of body motion generation, we compare the results of the same T2M-GPT model trained on both datasets. As detailed in Table~\ref{tab: cross-dataset}, the model trained on our dataset surpasses the one trained on the HumanML3D dataset on both test datasets. This outcome demonstrates that our additional datasets, namely Mixamo, GRAB, and IDEA400, provide more data whose distribution is not encompassed in the HumanML3D dataset, which greatly improved the motion diversity and model generalizability. This characteristic further solidifies the superiority of our model compared to previous models trained on the HumanML3D dataset.



\noindent\textbf{Whole-body Motion Results.}
For whole body motion evaluation, we focus on the motion consistency among different body parts. Table~\ref{tab:body-hand-matching} shows the results.
To generate the evaluation results, we train the feature extractor networks on respective dataset combinations for each modality pair.
The introduction of consistency loss in our model effectively improve the coherence of motion modalities. This observation is also validated in our qualitative results in Figure~\ref{fig:qualitative}. More examples are provided in the supplementary materials.

In summary, the outstanding performance of our model stems from our model's robust design, featuring VQ-VAE experts and a multi-indexing GPT for diverse data modalities with a modality alignment mechanism, and also the quality of our dataset with detailed text descriptions. Additional ablation studies are in the supplementary material.

\subsection{Qualitative Results}

\begin{figure}
    \centering
    \includegraphics[width=3in, height=2.4in]{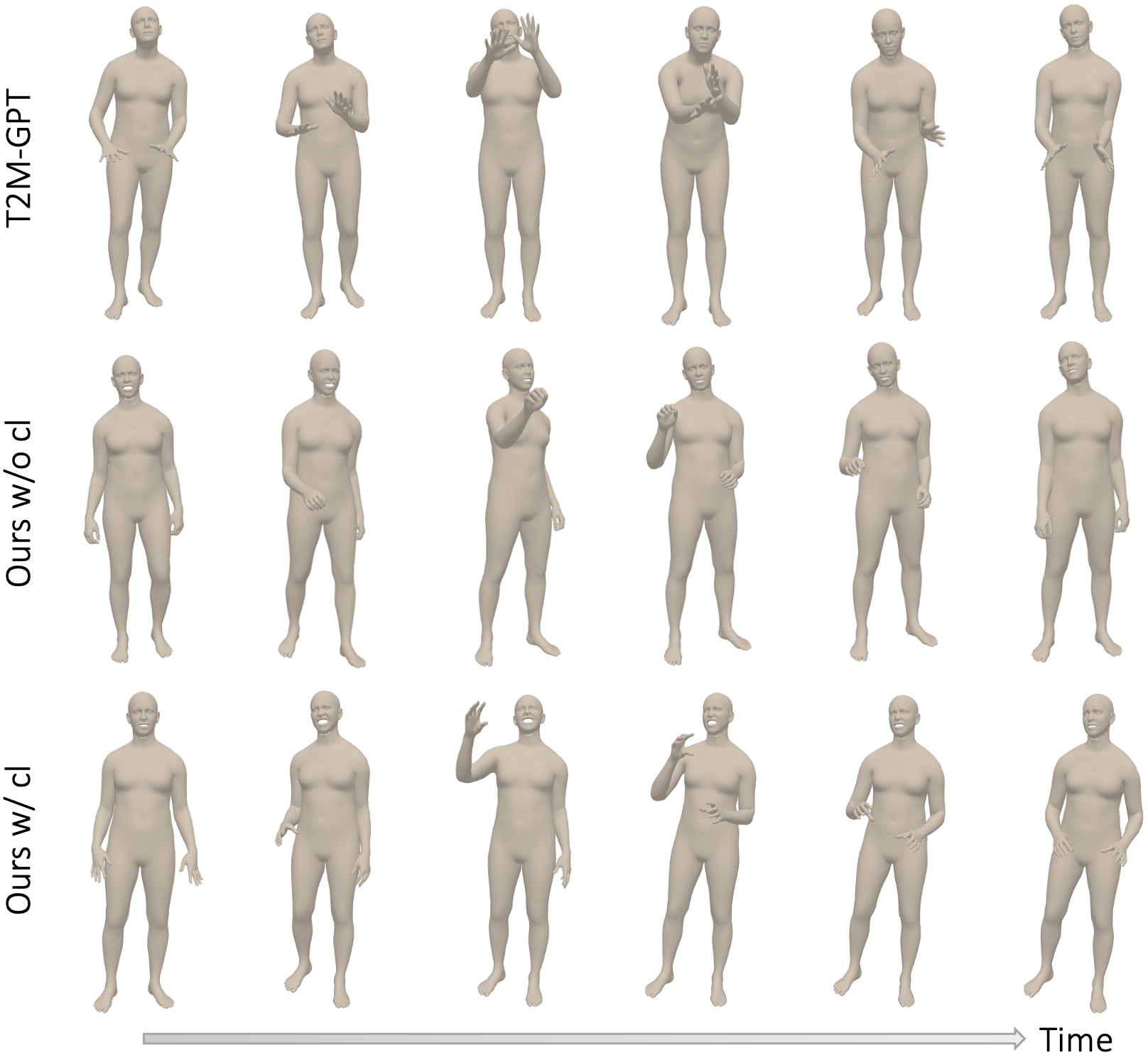}
    \caption{Qualitative results of the generated motions from our T2M-X models with and without consistency loss, compared with the T2M-GPT baseline.
    } 
    \vspace{-10pt}
    \label{fig:qualitative}
\end{figure}

Figure~\ref{fig:qualitative} illustrates qualitative results for three models: the T2M-GPT baseline, our model with/without consistency learning, using the example text input ``a person steps forward to pick up the fruit from the tree with a happy face''. 
The baseline lacks hand and face motion, making the output least expressive. 
Our model without consistency learning exhibits reasonable hand and face motions. However, the happy face smile comes at the very beginning in the motion sequence, which does not match the body motion very well. 
On the other hand, our model with consistency learning, displays the most clear and expressive body, hand, and face motion, with all modalities harmoniously matching. The face expression is also more expressive compared to the results without consistency learning. Due to the page limit, please refer to our supplementary materials for more visual examples.

\section{Conclusion}
In conclusion, we introduced T2M-X, a novel two-stage approach that learns expressive text-to-motion generation from partially annotated. It employs three distinct VQ-VAEs for the body, hands, and face, each trained on high-quality data sources. Furthermore, a Multi-indexing GPT model with a motion consistency loss function is used to generate and coordinate motions among different body parts. Our findings demonstrate that T2M-X significantly outperforms models trained for text to body motion generation task, both quantitatively and qualitatively.

{
    \small
    \bibliographystyle{ieeenat_fullname}
    \bibliography{main}
}


\end{document}

%% file: preamble.tex
\usepackage[dvipsnames]{xcolor}
